\documentclass[10pt,twocolumn,letterpaper]{article}

\usepackage[pagenumbers]{cvpr} % To force page numbers, e.g. for an arXiv version

\usepackage{colortbl}
\usepackage{graphicx}
\usepackage{subcaption}
\usepackage{xspace}
\usepackage{multirow}
\usepackage{makecell}
\usepackage{algorithm}
\newcommand{\bH}{\mathbf{H}}

\newcommand{\bn}{\mathbf{n}}\newcommand{\bN}{\mathbf{N}}

\newcommand{\bp}{\mathbf{p}}

\newcommand{\bR}{\mathbf{R}}
\newcommand{\bS}{\mathbf{S}}
\newcommand{\bt}{\mathbf{t}}
\newcommand{\bu}{\mathbf{u}}

\makeatletter
\DeclareRobustCommand\onedot{\futurelet\@let@token\@onedot}
\def\@onedot{\ifx\@let@token.\else.\null\fi\xspace}

\makeatother

\definecolor{yellow}{rgb}{1, 1, 0.7}
\definecolor{orange}{rgb}{1, 0.85, 0.7}
\definecolor{tablered}{rgb}{1, 0.7, 0.7}
\definecolor{red}{rgb}{1, 0, 0}
\definecolor{wincolor}{rgb}{0.85, 0.0, 0.0}
\definecolor{darkyellow}{rgb}{0.8, 0.8, 0.5}
\definecolor{darkred}{rgb}{0.7, 0.3, 0.3}
\definecolor{darkgreen}{rgb}{0.3, 0.7, 0.3}
\definecolor{green}{rgb}{0, 1.0, 0}
\definecolor{pink}{rgb}{1, 0.4, 0.7}

\definecolor{yzybest}{rgb}{1.0, 0.568, 0.584}
\definecolor{yzysecond}{rgb}{0.98, 0.78, 0.57}
\definecolor{yzythird}{rgb}{1.0, 1.0, 0.56}
\usepackage{times}
\usepackage{epsfig}
\usepackage{graphicx}
\usepackage{amsmath}
\usepackage{amssymb}
\usepackage{cuted}
\usepackage{amsfonts}
\usepackage{amsthm}
\usepackage{url}
\usepackage{color}
\usepackage{bbm}
\usepackage{tabularx}
\usepackage{xspace}
\usepackage{caption}
\usepackage{booktabs}
\usepackage{microtype}
\usepackage{wrapfig}
\usepackage{adjustbox}
\usepackage[outline]{contour}
\makeatletter
\@namedef{ver@everyshi.sty}{}
\makeatother
\usepackage{tikz}
\usepackage{bbm}
\usepackage{mathtools}
\usepackage{subcaption}
\usepackage{float}

\usepackage{multirow}
\definecolor{cvprblue}{rgb}{0.21,0.49,0.74}
\usepackage[pagebackref,breaklinks,colorlinks,allcolors=cvprblue]{hyperref}

\def\paperID{****} % *** Enter the Paper ID here
\def\confName{CVPR}
\def\confYear{2025}

\title{GlossGau: Efficient Inverse Rendering for Glossy Surface with \mbox{Anisotropic Spherical Gaussian}}

\author{Bang Du
~~~
Runfa Blark Li
~~~
Chen Du
~~~
Truong Nguyen\\
University of California San Diego\\
{\tt\small \{b7du,rul002,c9du,tqn001\}@ucsd.edu}
}

\begin{document}

\maketitle

\begin{strip}
    \centering
    \vspace{-5em}
    \begin{minipage}{\textwidth}
        \begin{figure}[H]
            \begin{subfigure}{0.7\textwidth}
                \includegraphics[width=\linewidth]{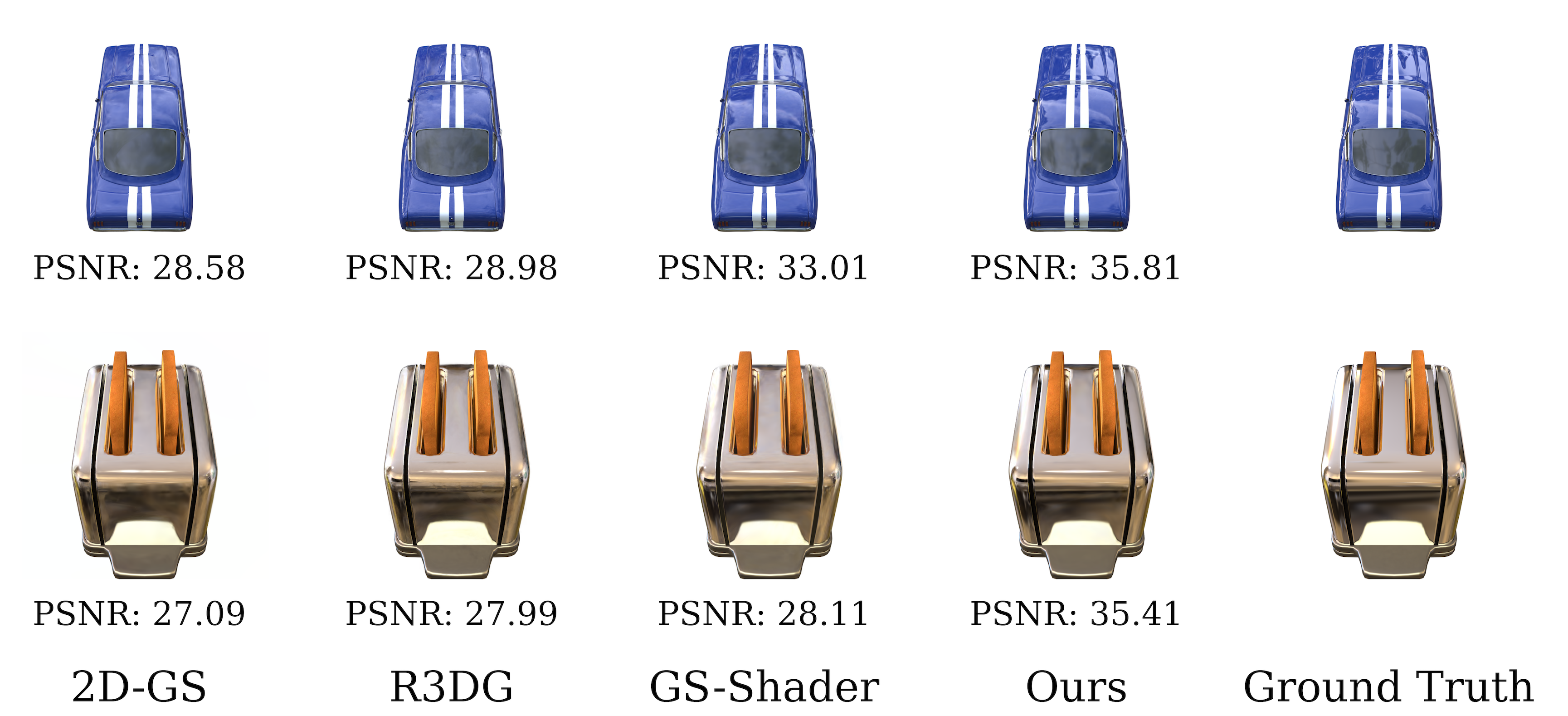}
                % \caption{} \label{fig:1a}
                % \vspace{3em}
            \end{subfigure}%
            \begin{subfigure}{0.3\textwidth}
                \includegraphics[width=\linewidth]{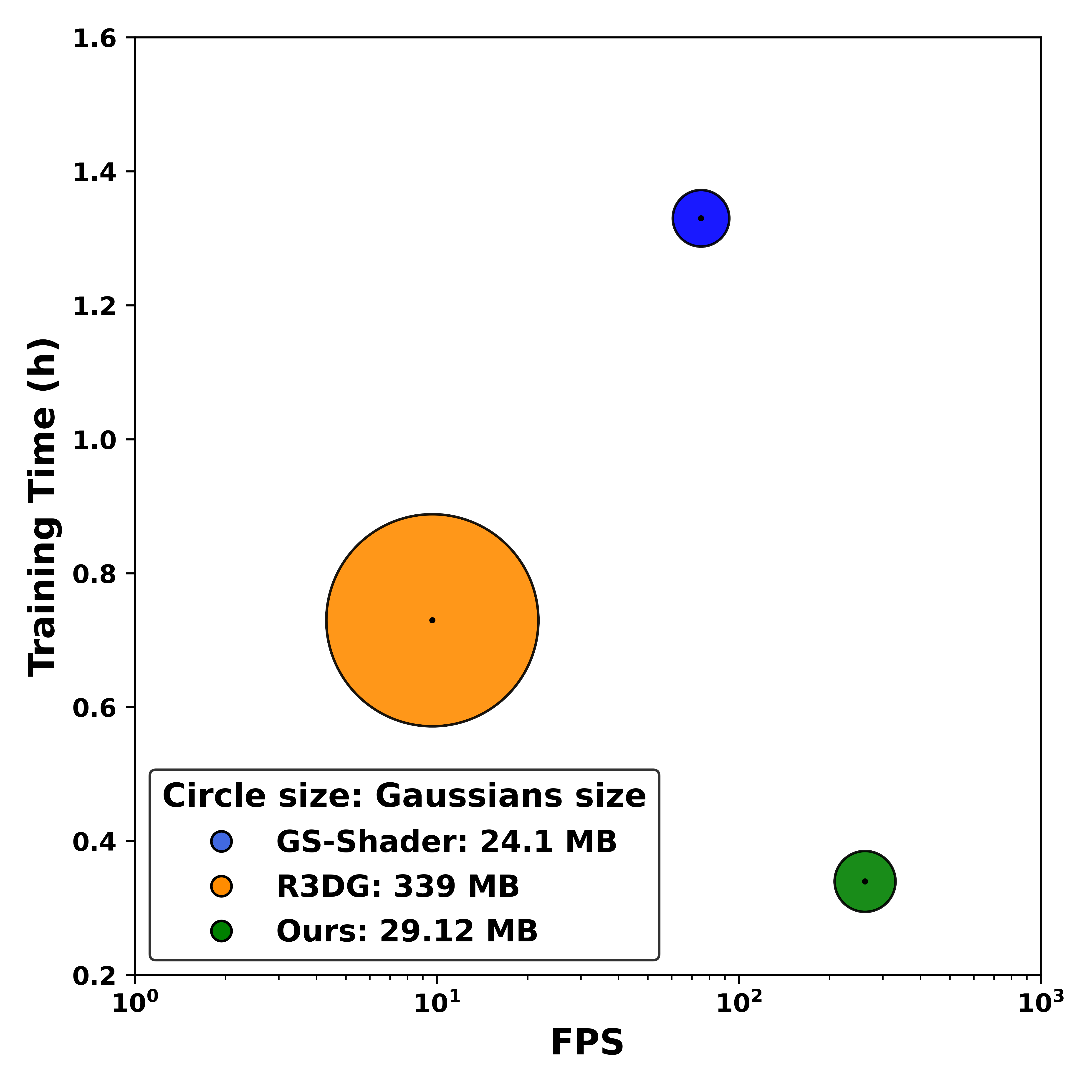}
                % \caption{} \label{fig:2a}
            \end{subfigure}%
            % \vspace{-5mm}
            \caption{GlossGau preserves the real-time rendering capabilities of 3D-GS while producing photorealistic images across diverse surface materials. Our method successfully captures glossy highlights and shadowing without compromising the quality of surrounding areas. GlossGau utilizes less storage space and optimizes at faster speed compared with \cite{jiang2024gaussianshader} and \cite{gao2023relightable}.}
            \label{fig:fig_teaser}
        \end{figure}
    \end{minipage}
\end{strip}

% \begin{strip}
%     \centering
%     \vspace{-5em}
%     \begin{subfigure}{0.7\textwidth}
%         \includegraphics[width=\linewidth]{figs/figure_cover_01.png}
%         \caption{} \label{fig:1a}
%     \end{subfigure}%
%     \begin{subfigure}{0.3\textwidth}
%         \includegraphics[width=\linewidth]{figs/figure_cover_02.png}
%         \caption{} \label{fig:2a}
%     \end{subfigure}%
%     \vspace{-5mm}
%     \captionof{figure}{
%     GlossGau preserves the real-time rendering capabilities of 3D-GS while producing photorealistic images across diverse surface materials. Our method successfully captures glossy highlights and shadowing without compromising the quality of surrounding areas. GlossGau utilizes less storage space and optimizes at faster speed compared with \cite{jiang2024gaussianshader} and \cite{gao2023relightable}.
%     }
%     \label{fig:fig_teaser}
%     % \vspace{-3mm}
% \end{strip}

\begin{abstract}

The reconstruction of 3D objects from calibrated photographs represents a fundamental yet intricate challenge in the domains of computer graphics and vision. Although neural reconstruction approaches based on Neural Radiance Fields (NeRF) have shown remarkable capabilities, their processing costs remain substantial. Recently, the advent of 3D Gaussian Splatting (3D-GS) largely improves the training efficiency and facilitates to generate realistic rendering in real-time. However, due to the limited ability of Spherical Harmonics (SH) to represent high-frequency information, 3D-GS falls short in reconstructing glossy objects. Researchers have turned to enhance the specular expressiveness of 3D-GS through inverse rendering. Yet these methods often struggle to maintain the training and rendering efficiency, undermining the benefits of Gaussian Splatting techniques. In this paper, we introduce GlossGau, an efficient inverse rendering framework that reconstructs scenes with glossy surfaces while maintaining training and rendering speeds comparable to vanilla 3D-GS. Specifically, we explicitly model the surface normals, Bidirectional Reflectance Distribution Function (BRDF) parameters, as well as incident lights and use Anisotropic Spherical Gaussian (ASG) to approximate the per-Gaussian Normal Distribution Function under the microfacet model. We utilize 2D Gaussian Splatting (2D-GS) as foundational primitives and apply regularization to significantly alleviate the normal estimation challenge encountered in related works. Experiments demonstrate that GlossGau achieves competitive or superior reconstruction on datasets with glossy surfaces. Compared with previous GS-based works that address the specular surface, our optimization time is considerably less. 

\end{abstract}    
\section{Introduction}
\label{sec:intro}

Photorealisric reconstruction from images captured from multiple viewpoints is a critical yet challenging task in computer graphics and computer vision. In recent years, the field of 3D computer vision has witnessed remarkable advancements in the 3D reconstruction and visualization of 3D scenes. Innovations such as Neural Radiance Fields (NeRF)~\cite{mildenhall2020nerf} have achieved substantial breakthroughs in generating novel views of 3D objects and scenes, presenting the potential for high-quality and photo-realistic renderings. More recently, 3D Gaussian Splatting~\cite{kerbl20233d} combines 3D Gaussian representation and tile-based splatting techniques to achieve high-quality 3D scene modeling and real-time rendering. Despite its exceptional performance, 3D-GS struggles to model glossy surfaces within scenes. 3D Gaussian Splatting~\cite{kerbl20233d} utilizes low-order spherical harmonics (SH) instead of explicitly model appearance properties, so that fails to capture significant view-dependent changes, particularly specular highlights. Recent studies~\cite{liang2024gs,gao2023relightable,jiang2024gaussianshader} try to build inverse rendering pipelines based on 3D-GS. However, these methods mostly sacrifices training and rendering efficiency, thereby undermining the benefits of Gaussian Splatting techniques.

In this paper, we introduce GlossGau, an efficient inverse rendering framework that reconstructs scenes with glossy surfaces while preserving computational performance of 3D-GS. Our approach jointly optimizes illumination, material properties, geometry, and surface normals from multi-view images with known camera poses. To maintain computational efficiency, we formulate specular BRDF terms using anisotropic spherical Gaussian. For normal estimation and geometry consistency, we utilize the surfel-based Gaussian Splatting, which collapses volumetric representations into thin surfaces, to rigorously define the geometry of Gaussians. The resulting Gaussians reconstruct glossy surfaces at fast training and rendering speed and supports post-manipulations such as material editing and relighting. In summary, our key contributions are:
\begin{enumerate}
    \item Our method explicitly approximates the rendering equation and analytically represents the specular BRDF using Anisotropic Spherical Gaussian, yielding photorealistic reconstruction particularly for glossy surfaces compared to standard 3D-GS, without compromising the optimization efficiency. 
    \item Leveraging surfel-based Gaussian Splatting and regularization, our approach achieves enhanced precision in surface normal estimation, reducing the material-geometry ambiguities and resulting in better inverse rendering. 
    \item Extensive experimental evaluation across diverse datasets validates GlossGau's high rendering fidelity for both general and highly-glossy surfaces. Maintaining comparable visual quality, our approach achieves up to 66\% reduction in optimization time and supports real-time rendering performance compared to existing GS-based inverse rendering methods.
\end{enumerate}

\begin{figure*}[t]
    \centering
    \includegraphics[width=\linewidth]{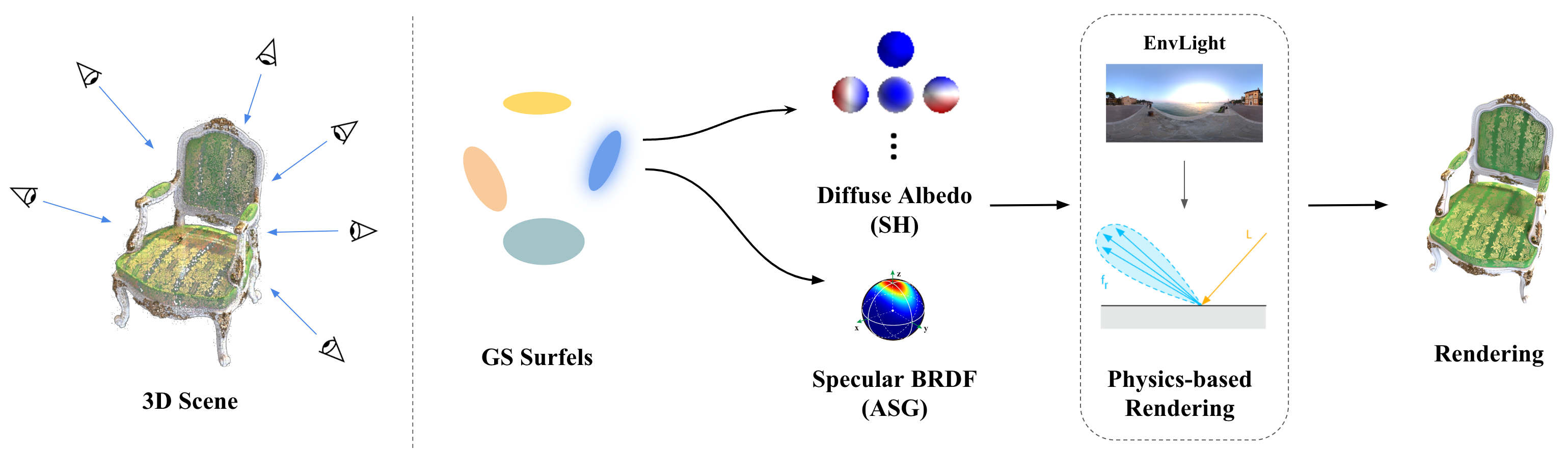}
    \vspace{-3ex}
    \caption{\textbf{GlossGau Pipeline.} Our framework initializes with surfel-based Gaussian primitives, where each primitive carries both geometric attributes (position, covariance, and opacity) and appearance properties. The appearance representation decomposes into view-dependent diffuse albedo, parameterized via Spherical Harmonics, and specular BRDF terms, modeled through Anisotropic Spherical Gaussians. The final rendering integrates these material maps with differentiable environment lighting to achieve high-fidelity inverse rendering results.
    }
    \label{fig:pipe}
    \vspace{-2ex}
\end{figure*}

\section{Related Works}
\label{sec:related}
\noindent{\bf Neural Radiance Field.}
Neural Radiance Fields (NeRF) \cite{mildenhall2020nerf} have achieved remarkable progress in photo-realistic novel view synthesis through implicit representations and volume rendering. Subsequent research has expanded NeRF to various applications. \cite{barron2021mip, barron2022mip} which enhance rendering fidelity through 3D conical frustum sampling, establishing new benchmarks in novel view synthesis. Several works \cite{oechsle2021unisurf, yariv2021volume, wang2021neus, yu2022monosdf} augment NeRF with implicit surface representations to achieve higher geometric accuracy. Some studies \cite{srinivasan2021nerv, zhang2021nerfactor, yao2022neilf, verbin2022ref} address the challenges of modeling specular surfaces and view-dependent reflections. A parallel direction of research \cite{muller2022instant, fridovich2022plenoxels, liu2020neural, chen2022tensorf, garbin2021fastnerf, sun2022direct} tackles NeRF's computational limitations through spatial acceleration structures such as voxel grids and hash encodings. However, despite these innovations, NeRF-based methods  continue to face constraints in rendering performance and memory efficiency during training.

\noindent{\bf 3D Gaussian Splatting.}
Point primitives as a fundamental rendering element were first proposed by~\cite{levoy1985use}. Following seminal developments in point-based rendering~\cite{aliev2020neural,xu2022point,zhang2022differentiable}, 3D Gaussian Splatting~\cite{kerbl20233d} represents a breakthrough that enables real-time photorealistic neural rendering through efficient tile-based rasterization of splatted 3D Gaussian primitives. This methodology bypasses the computational burden of dense ray sampling while surpassing implicit neural representations in both visual fidelity and computational efficiency. Subsequently, various enhancements have emerged to improve 3D-GS. \cite{Yu2023MipSplatting} utilizes a 3D smoothing filter and a 2D mip filter to reduce aliasing when zooming in and zooming out. \cite{huang2024erroranalysis3dgaussian} proposes an optimal projection strategy by minimizing projection errors, resulting in photo-realistic rendering under various focal lengths and various camera models. Scaffold-GS~\cite{lu2024scaffold} multi-scale hybrid architecture that combines explicit 3D Gaussians with implicit MLP networks, significantly reducing the size of 3D-GS and improving the rendering quality. \cite{Huang2DGS2024} extends it to a perspective-accurate 2D splatting process for detailed geometry.

\noindent{\bf Material and Lighting Estimation.}
Reconstruction and inverse rendering of a glossy object from multview images has been a challenging task due to the complex light interactions and ambiguities. Some methods simplify the problem by assuming controllable lighting conditions~\cite{azinovic2019inverse,guo2019relightables,park2020seeing,bi2020DRV,bi2020deep3d,nam2018practical,schmitt2020joint,bi2020NRF}. Subsequent research has explored more complex lighting models to handle realistic scenarios. NeRV~\cite{srinivasan2021nerv} and PhySG~\cite{zhang2021physg} leverage environmental maps to manage arbitrary lighting conditions. NeRD~\cite{boss2021nerd} addresses the challenge of images captured under varying illumination by attributing Spherical Gaussians to each image.  Ref-NeRF~\cite{verbin2022ref} employs integrated direction encoding to improve the reconstruction fidelity of reflective objects. NeRO~\cite{liu2023nero} use Integrated Directional Encoding similar to Ref-NeRF~\cite{verbin2022ref} and employs a two-stage inverse rendering strategy. TensoIR~\cite{jin2023tensoir} employs TensoRF~\cite{chen2022tensorf} as its geometry representation and utilizes ray marching to compute indirect illumination and visibility. However, these methods that rely on implicit representations~\cite{zhang2021physg, verbin2022ref, liu2023nero, karis2013real, jin2023tensoir, srinivasan2021nerv, wu2023nerf, zeng2023relighting} involve densely sampled points along rays, leading to a significant decrease in the performance of inverse rendering.

Recent works~\cite{jiang2024gaussianshader,liang2024gs, gao2023relightable} aim to employ 3D-GS~\cite{kerbl20233d} as an explicit representation of scenes in order to enhance the performance of inverse rendering. GaussianShader~\cite{jiang2024gaussianshader} decomposes the original color attributes into components including diffuse, specular and residual color for modeling the color of reflective surfaces. In GS-IR~\cite{liang2024gs}, a technique similar to light probes~\cite{debevec2008median} is employed to obtain the indirect light and ambient occlusion. These methods~\cite{jiang2024gaussianshader, liang2024gs} encounter challenges in representing accurate surfaces and materials using 3D Gaussians and struggle to disambiguate between macroscopic surface normals and microscopic roughness parameters, leading to reduced inverse rendering quality and extended optimization times. \cite{gao2023relightable} directly estimates BRDF parameters for each Gaussian and utilizes the normal-based densification to ensure the volumetric rendering quality. For complex surfaces, its result suffers from over-inflated size and significantly reduces rendering speed. In contrast, our method effectively preserves the efficiency and achieves comparable rendering quality with better geometry estimation. Our method effectively balances the inverse rendering quality and the space and time complexity of optimization.

\newcommand{\viewdir}{{\boldsymbol{\omega}_o}}
\newcommand{\lightdir}{{\boldsymbol{\omega}_i}}
\newcommand{\location}{{\mathbf{x}}}
\newcommand{\normaldir}{{\mathbf{n}}}

\newcommand{\lightsgsharp}{\lambda}
\newcommand{\lightsgdir}{\boldsymbol{\xi}}
\newcommand{\lightsgamp}{\boldsymbol{\mu}}

\newcommand{\brdfsgsharp}{\lambda}
\newcommand{\brdfsgdir}{\boldsymbol{\xi}}
\newcommand{\brdfsgamp}{\boldsymbol{\mu}}

\newenvironment{packed_itemize}{
\begin{list}{\labelitemi}{\leftmargin=2em}
\vspace{-6pt}
 \setlength{\itemsep}{0pt}
 \setlength{\parskip}{0pt}
 \setlength{\parsep}{0pt}
}{\end{list}}

\section{Method}

In the following section, we describe the GlossGau pipeline. Our method begins with reconstructing Gaussian Surfels and estimates BRDF parameters for specular surface representation. We employs Anisotropic Spherical Gaussian to represent normal distribution function, which enhances the representation ability without additional optimization overhead. Moreover, we compute the visibility based upon the modified point-based ray tracing~\cite{gao2023relightable}, which contributes to the rendering of realistic shadow effect. The overview of our method is illustrated in Figure \ref{fig:pipe}.

\subsection{Preliminaries}
\textbf{3D Gaussian Splatting.} 3DGS~\cite{kerbl20233d} is a point-based method that employs explicit 3D Gaussian points as its rendering primitives. A 3D Gaussian point is mathematically defined by a center position $\boldsymbol{x}$, opacity $\sigma$, and a 3D covariance matrix $\Sigma$. The appearance $\boldsymbol{c}$ of each 3D Gaussian is represented using the first three orders of spherical harmonics (SH) to achieve the view-dependent variation. The 3D Gaussian value at point $x$ is evaluated as
\begin{equation}
G(\boldsymbol{x}) = \exp(-\frac{1}{2}(\boldsymbol{x}-\boldsymbol{\mu})^\top\Sigma^{-1}(\boldsymbol{x}-\boldsymbol{\mu}))
\label{eq:3d_gaussian}
\end{equation}
In practice, the 3D covariance matrix $\Sigma$ is decomposed into the quaternion $\boldsymbol{r}$, representing rotation, and the 3D vector $\boldsymbol{s}$, representing scaling such that $\Sigma = \boldsymbol{R}\boldsymbol{S}\boldsymbol{S^T}\boldsymbol{R^T}$ for better optimization. 

During the rendering process, 3D Gaussians are firstly ``splatted'' into 2D Gaussians on the image plane. The 2D means are determined by accurate projection of 3D means, while the 2D covariance matrices are approximated by:
$\Sigma'=\boldsymbol{J}\boldsymbol{W}\Sigma\boldsymbol{W}^\top\boldsymbol{J}^\top$,
where $\boldsymbol{W}$ and $\boldsymbol{J}$ represent the viewing transformation and the Jacobian of the affine approximation of perspective projection transformation~\cite{zwicker2001surface}. The pixel color is thus computed by alpha blending corresponding $N$ 2D Gaussians sorted by depth:
\begin{equation}
C(\textbf{p}) = \sum_{i\in{N}}T_{i}\alpha_{i}\boldsymbol{c}_{i}, \hspace{0.5em} T_i = \prod_{j=1}^{i-1}(1-\alpha_{j}) .
    \label{eq:alpha_blending}
\end{equation}
$\alpha$ is obtained by querying 2D splatted Gaussian associated with pixel $\textbf{p}$. 
\begin{equation}
    \alpha_i = \sigma_i e^{-\frac{1}{2} (\textbf{p} - \mu_i)^T \Sigma^{-1} (\textbf{p} - \mu_i) }
\end{equation}

\noindent\textbf{Surfel-based Gaussian Splatting.}
3D-GS is known to fall short in capturing intricate geometry because the volumetric 3D Gaussian, which models complete angular radiance, conflicts with the thin nature of surfaces. To achieve accurate geometry as well as high-quality novel view synthesis, methods \cite{Huang2DGS2024, dai2024high} simplify the 3-dimensional modeling by adopting Gaussian surfels embedded in 3D space. This 2D surfel approach allows each primitive to distribute densities across a planar disk, with the normal vector simultaneously defined as the direction exhibiting the steepest density gradient. In GlossGau, we follow the design of 2D-GS \cite{Huang2DGS2024} to reformulate the covariance matrix into two principal tangential vectors $\bt_u$ and $\bt_v$, and a scaling vector $\bS = (s_u,s_v)$. The scaling vector $s_z$ is enforced to be zero.

A 2D Gaussian is therefore defined in a local tangent plane in world space, which is parameterized:
\begin{gather}
    P(u,v) = \bp + s_u \bt_u u + s_v \bt_v v = \bH(u,v,1,1)^{\mathrm{T}}\\
    \text{where} \, \bH = 
    \begin{bmatrix}
        s_u \bt_u & s_v \bt_v & \boldsymbol{0} & \bp \\
        0 & 0 & 0 & 1 \\
    \end{bmatrix} = \begin{bmatrix}
        \bR\bS & \bp_k \\ 
        \boldsymbol{0} & 1\\
    \end{bmatrix}
    \label{eq:plane-to-world}
\end{gather}
in which $\bH \in {4\times 4}$ is a homogeneous transformation matrix representing the geometry of the Gaussian surfel. For the point $\bu=(u,v)$ in the local $uv$ space of the primitive, its 2D Gaussian value can then be evaluated by standard Gaussian
\begin{equation}
G(\bu) = \exp\left(-\frac{u^2+v^2}{2}\right)
    \label{eq:gaussian-2d}
\end{equation} 

\subsection{Glossy Appearance modeling}
Both 3D-GS and 2D-GS model the view-dependent appearances with low-order spherical harmonic functions, which has limited ability to capture the high-frequency information, such as specular surfaces in the scene. We model the light-surface interactions in a way consistent with the rendering equation~\cite{kajiya1986rendering}: 

\begin{equation}
L_{o}(\boldsymbol{\omega_{o}}, \boldsymbol{x}) = \int_{\Omega}f(\boldsymbol{\omega_{o}}, \boldsymbol{\omega_{i}}, \boldsymbol{x})L_{i}(\boldsymbol{\omega_{i}}, \boldsymbol{x})(\boldsymbol{\omega_{i}}\cdot\boldsymbol{n})d\boldsymbol{\omega_{i}}
\label{eq:rendering_equation}
\end{equation}
where $\boldsymbol{x}$ and $\boldsymbol{n}$ are the surface point and its normal vector, $f$ is the Bidirectional Reflectance Distribution Function (BRDF), and $L_{i}$ and $L_{o}$ denote the incoming and outgoing radiance in directions $\boldsymbol{\omega_{i}}$ and $\boldsymbol{\omega_{o}}$. $\Omega$ signifies the hemispherical domain above the surface. However, accurately modeling light-surface interactions necessitates a precise evaluation of the Rendering Equation, which can be time-consuming for point-based rendering and thus undermines the efficiency benefits offered by Gaussian Splatting. In our approach, the incident light $L_i$ is decomposed into a global 16x32 direct environment map and a local per-Gaussian indirect illumination component, parameterized using three-level SH. For each 3D Gaussian, we sample $N_s$ incident directions over the hemisphere space through Fibonacci sampling~\cite{yao2022neilf, gao2023relightable}, we have found to effectively balance efficiency during both training and rendering phases.

We further decompose the rendered color into diffuse and specular components: 
\begin{equation}
    \boldsymbol{c} = \boldsymbol{c}_d + \boldsymbol{c}_s 
\end{equation}
Since SH functions model low-frequency information in a low cost manner, only $c_s$, as the specular component, involves the computation of BRDF parameter for the best efficiency. We adopt the simplified Disney Model~\cite{burley2012physically,karis2013real} as in prior work~\cite{zhang2021physg,gao2023relightable}. The specular term of BRDF is defined as 
\begin{equation}
    f_{s}(\boldsymbol{\omega}_{o}, \boldsymbol{\omega}_{i})=\frac{D(\boldsymbol{h}; r) \cdot F(\boldsymbol{\omega}_{o},\boldsymbol{h}) \cdot G(\boldsymbol{\omega}_{i},\boldsymbol{\omega}_{o}, h; r)}{4(\boldsymbol{n} \cdot \boldsymbol{\omega}_{i}) \cdot (\boldsymbol{n} \cdot \boldsymbol{\omega}_{o})}
    \label{eq:specular_term}
\end{equation}
where $r\in[0, 1]$ is the surface roughness, $\boldsymbol{h}=({\boldsymbol{\omega}_{i}+\boldsymbol{\omega}_{o}}) / \|\viewdir + \lightdir\|_2$ is the half vector, \textit{D}, \textit{F} and \textit{G} denote the normal distribution function, Fresnel term and geometry term. We utilize isotropic Spherical Gaussians (SG)~\cite{wang2009all} to approximate \textit{D} under the microfacet-based model \cite{cook1981reflectance}. An $n$-dimensional SG is a spherical function that takes the form:
\begin{equation}\label{eq:sg} 
    G(\boldsymbol{\nu}; \lightsgdir, \lightsgsharp, \lightsgamp) = \lightsgamp \, e^{\lightsgsharp (\boldsymbol{\nu} \cdot \lightsgdir - 1)}
\end{equation}
where $\boldsymbol{\nu} \in \mathbb{S}^2$ is the function input, $\lightsgdir \in \mathbb{S}^2$ is the lobe axis, $\lightsgsharp \in \mathbb{R}_+$ is the lobe sharpness, and $\lightsgamp \in \mathbb{R}_+^n$ is the lobe amplitude. Under the isotropic assumption, we have the approximated \textit{D} after normalization as
\begin{equation}
    D(\boldsymbol{h}) = G(\boldsymbol{h}; \boldsymbol{n}, \frac{2}{r^2}, \frac{1}{\pi r^2})
\end{equation}

However, to faithfully represent real-world BRDFs, which typically exhibit some degree of anisotropy, \cite{wang2009all} relies on a mixture of $n$ distributed Spherical Gaussians. While effective, this mixture model introduces optimization overhead. Anisotropic spherical Gaussian (ASG) \cite{xu2013anisotropic} extends SG and has been demonstrated to effectively model anisotropic scenes with a small number of functions. The ASG function is defined as:
\begin{equation}
\label{equ: asg-function}
\begin{aligned}
    ASG(\mathbf{\nu} \: | \: [\mathbf{x}, \mathbf{y}, \mathbf{z}],[\lambda, \mu], c)= c \cdot \mathrm{S}(\mathbf{\nu} ; \mathbf{z}) \cdot e^{-\lambda(\mathbf{\nu} \cdot \mathbf{x})^{2}-\mu(\mathbf{\nu} \cdot \mathbf{y})^{2}}
\end{aligned}
\end{equation}
where $\nu$ is the unit direction serving as the function input; $\mathbf{x}$, $\mathbf{y}$, and $\mathbf{z}$ correspond to the tangent, bi-tangent, and lobe axis, respectively, and are mutually orthogonal; $\lambda$ and $\mu$ are the sharpness parameters for the $\mathbf{x}$- and $\mathbf{y}$-axis, satisfying $\lambda, \mu>0$; $c$ is the lobe amplitude; $\mathrm{S}$ is the smooth term defined as $\mathrm{S}(\nu ; \mathbf{z}) = \max (\nu \cdot \mathbf{z}, 0)$. 
At the rendering stage, we apply the warping operator presented in \cite{xu2013anisotropic} that takes a normal distribution function approximated as a SG and stretch it along the view direction to produce an ASG that better represents the actual BRDF. We thereby obtain a more accurate BRDF representation while eliminating the need for additional parameter estimation. More detailed are provided in the supplementary. 

In summary, the ASG parameters of 2D Gaussians remain determined exclusively by two factors: the surface normal vector $\boldsymbol{n}$ and roughness value $r$. The next critical step involves calculating view-dependent visibility, which we detail in the following section.

\noindent\textbf{Visibility Approximation.}
Computing visibility values through online ray tracing operations presents significant performance bottlenecks due to its intensive computational demands. Inspired by \cite{gao2023relightable,moenne20243d}, we implement the ray tracing for 2D-GS and precompute the visibility term \textit{V}. Since 2D-GS primitives are primarily thin surfaces, naively constructing a Bounding Volume Hierarchy (BVH) using axis-aligned bounds as AABBs significantly increases the false-positive intersections during the traversal, resulting needless computations. We thereby construct the bounding proxy as an anisotropically scaled icosahedron~\cite{moenne20243d} with a threshold response value $\alpha_{min}$. The scaling factor takes into account particle opacity, allowing for more efficient bounds. We formulate a precise mathematical framework for computing Gaussian particle responses in 3D ray tracing. Rather than a segment falls within a 3D volume, our ray-surfel intersection reduces to a point and the response can be directly evaluated using Eq. \ref{eq:gaussian-2d}. This formulation eliminates the need for approximate solutions previously employed to resolve intersection ambiguities. Furthermore, in contrast to prior approaches, the inverse of our covariance matrix $\bH$ is analytically formulated, which significantly enhances numerical stability of computing particle response on flat Gaussian primitives. We provide detailed information in the supplementary.

\subsection{Regularizations}
\label{sec:regularization}

\noindent{\bf Surface Normal Estimation.}
Accurate geometric representation is essential for realistic physically-based rendering, as emphasized by the formulation of rendering equation. Predicting accurate normals on discrete Gaussian spheres has been a recognized challenge~\cite{kerbl20233d,Huang2DGS2024,jiang2024gaussianshader,liang2024gs,gao2023relightable}. Although adapting 2D Gaussian methodology demonstrates strong performance in geometric representation, relying exclusively on photometric loss functions can result in reconstruction artifacts. We incorporate a normal attribute $\bn$ into the set of optimizable attributes for each 3D Gaussian. Instead of selecting the shortest axis and predicting residuals for better alignment in 3D-GS, we are able to explicitly define the normal direction as the cross product of two tangential vectors that define the Gaussian surfel. This normal is initialized in association with the Gaussian and optimized through the differentiable rasterization process alongside the tangential vectors. $\bn$ is supervised by the consistency between the rendered normal map and depth gradient, enforcing geometric alignment between the surface normals and depth-derived geometry.

\begin{equation}
\mathcal{L}_{n} = \sum_{i} \omega_i (1-\bn_i^\mathrm{T}\bN)
\vspace{-1ex}
\end{equation}
where $i$ indexes over intersected splats along the ray, $\omega$ denotes the blending weight of the intersection point, $\bn_i$ represents the normal of the splat that is oriented towards the camera, and $\bN$ is the normal estimated by the gradient of the depth map. Specifically, $\bN$ is computed with finite differences from nearby depth points as follows:
\begin{equation}
\mathbf{N}(x,y) = \frac{\nabla_x \bp_s \times \nabla_y \bp_s}{|\nabla_x \bp_s \times \nabla_y \bp_s|}
\label{eq:normal_depth}
\end{equation} 
By orienting each Gaussian splat's normal vector parallel to the estimated surface normal, we ensure the 2D projections locally conform to the underlying geometric surface.

\noindent{\bf Light Regularization.}
To tackle the inherent material-illumination ambiguity, we apply physics-based regularization terms~\cite{gao2023relightable} that guide the separation of surface properties from lighting effects. Following the assumption of approximately neutral incident illumination~\cite{liu2023nero}:
\begin{equation}
\label{eq:reg_light}
    \mathcal{L}_{light} = \sum\nolimits_{c}(L_{c} - \frac{1}{3}\sum\nolimits_{c}L_{c}), c\in\{R,G,B\}.
\end{equation}

\noindent{\bf Alpha Mask Constraint. }
We observe the optimization of 2D-GS on bounded scenes sometimes can be unstable under normal regularization, where the background Gaussians cannot be well eliminated. When the foreground mask is provided, we constrain the training by L1 loss. 
\begin{equation}
    \mathcal{L}_{alpha} = \| M - O\|
\end{equation}
where $M$ is the alpha channel of the input image and $O$ is the rendered opacity.

\section{Experiments}
\label{sec:experiments}
\subsection{Training Details}
\label{sec:details}

\begin{table*}
\centering
\vspace{-2ex}
\caption{Quantitative comparisons on NeRF Synthetic dataset. We report color each cell as \colorbox{yzybest}{best}, \colorbox{yzysecond}{second best} and \colorbox{yzythird}{third best}.}
\vspace{-1ex}
\scalebox{0.85}{
\begin{tabular}{lccccccccc}
\hline
        \multicolumn{10}{c}{NeRF Synthetic~\cite{mildenhall2020nerf}}       \\
            & \multicolumn{1}{l}{Chair} & \multicolumn{1}{l}{Drums} & \multicolumn{1}{l}{Lego} & \multicolumn{1}{l}{Mic} & \multicolumn{1}{l}{Materials} & \multicolumn{1}{l}{Ship} & \multicolumn{1}{l}{Hotdog} & \multicolumn{1}{l}{Ficus} & \multicolumn{1}{l}{Avg.}           \\ \hline
\multicolumn{10}{c}{PSNR$\uparrow$}                                                                                                                                                                                                                                               \\ \hline
2D-GS~\cite{Huang2DGS2024}        & \cellcolor{yzythird}{34.89}                     & \cellcolor{yzythird}{26.02}                     & 33.61        & 33.99                   & \cellcolor{yzythird}{30.15}                        & \cellcolor{yzythird}{30.10}                    & {36.84}                      & \cellcolor{yzysecond}{35.79}   &  \cellcolor{yzythird}32.67                          \\

Scaffold-GS~\cite{lu2024scaffold}      & \cellcolor{yzythird}{34.89}                     & \cellcolor{yzysecond}{26.26}                     & \cellcolor{yzysecond}35.05                    & \cellcolor{yzybest}{36.23}                  & \cellcolor{yzysecond}30.30                         & \cellcolor{yzybest}30.10                    & \cellcolor{yzybest}37.70        & \cellcolor{yzythird}34.39 & \cellcolor{yzysecond}33.12                             \\

GS-Shader~\cite{jiang2024gaussianshader}    & 33.36                     & 25.52                     & 33.05                   & 34.06                   & 28.82                        & 28.37                   & 35.16                      & 33.04    &  31.42             \\

GS-IR~\cite{liang2024gs}    & 28.71                     & 24.44                     & 32.49                   & 31.95               & 26.62                        & 27.96                   & 34.23                      & 30.15    &  29.57             \\

R3DG~\cite{gao2023relightable}    &  \cellcolor{yzysecond}{34.90}                    & 24.06                    &   \cellcolor{yzybest}{35.55}                 &  \cellcolor{yzysecond}{35.25}                  &   29.54                      &  28.75                   &    \cellcolor{yzythird}36.96                  &   34.66  &  32.46            \\

Ours & \cellcolor{yzybest}{35.51}                          & \cellcolor{yzybest}{26.54}            & \cellcolor{yzythird}{34.67}                         & \cellcolor{yzythird}{35.12}                        & \cellcolor{yzybest}{30.79}         & \cellcolor{yzysecond}{30.60}                       & \cellcolor{yzysecond}{37.63}                           & \cellcolor{yzybest}{36.45}  &  \cellcolor{yzybest}{33.41}                          \\ \hline

\multicolumn{10}{c}{SSIM$\uparrow$}                                                                                                                                                                                                                                            \\ \hline
2D-GS~\cite{Huang2DGS2024}        & \cellcolor{yzybest}0.987               & \cellcolor{yzybest}0.953   & \cellcolor{yzythird}0.977                    & \cellcolor{yzythird}0.990                   & \cellcolor{yzybest}0.958              & \cellcolor{yzybest}0.904                    & \cellcolor{yzysecond}0.983                      & \cellcolor{yzybest}0.987   &  \cellcolor{yzybest}0.967                        \\

Scaffold-GS~\cite{lu2024scaffold}      & \cellcolor{yzysecond}0.984                     & \cellcolor{yzythird}0.948                     & \cellcolor{yzysecond}0.980                    & \cellcolor{yzythird}0.990                  & \cellcolor{yzybest}0.958                         & \cellcolor{yzysecond}0.898                    & \cellcolor{yzysecond}0.983                      & \cellcolor{yzythird}0.985 & \cellcolor{yzysecond}0.966                             \\

GS-Shader~\cite{jiang2024gaussianshader}    & 0.979                     & 0.945                    & 0.972                    & 0.988                   & 0.951                         & \cellcolor{yzythird}0.881                   & 0.979                      & 0.982    &  0.960                        \\

GS-IR~\cite{liang2024gs} & 0.954 & 0.920 & 0.963 & 0.968 & 0.914 & 0.858 & 0.969 & 0.958   & 0.938     \\

R3DG~\cite{gao2023relightable}    &  \cellcolor{yzythird}0.982            & \cellcolor{yzysecond}0.952       & \cellcolor{yzybest}0.982                 &  \cellcolor{yzybest}0.992                  &   \cellcolor{yzythird}0.955                      &  0.878                   &    \cellcolor{yzysecond}0.984               &   \cellcolor{yzysecond}0.986  &  0.964                                                                                                                                                                                                                                                                     \\

Ours & \cellcolor{yzybest}{0.988}                          & \cellcolor{yzybest}{0.953}              & \cellcolor{yzybest}{0.982}                         & \cellcolor{yzysecond}{0.991}                        & \cellcolor{yzysecond}{0.956}             & {0.879}                       & \cellcolor{yzybest}{0.985}                           & \cellcolor{yzybest}{0.987}  &  \cellcolor{yzythird}{0.965}              
 \\ \hline

\multicolumn{10}{c}{LPIPS$\downarrow$}                                                                                                                                                                                                                                              \\ \hline
2D-GS~\cite{Huang2DGS2024}        & \cellcolor{yzybest}0.010                     & \cellcolor{yzysecond}0.040                     & 0.022                   & \cellcolor{yzysecond}0.007                   & \cellcolor{yzythird}0.038                         & \cellcolor{yzythird}0.117                    & 0.024                      & \cellcolor{yzysecond}0.012  &  \cellcolor{yzythird}0.034                          \\

Scaffold-GS~\cite{lu2024scaffold}      & \cellcolor{yzythird}0.014                     & 0.047                     & \cellcolor{yzythird}0.019                 & \cellcolor{yzythird}0.008                  & 0.042                         & \cellcolor{yzybest}0.111                    & \cellcolor{yzythird}0.023                     & 0.014 & 0.035                  \\

GS-Shader~\cite{jiang2024gaussianshader}    & 0.019                     & \cellcolor{yzythird}0.045                     & 0.026             &0.009                   & 0.046                         & 0.147                   & 0.030                      & 0.017    &  0.042                       \\

GS-IR~\cite{liang2024gs} & 0.038 & 0.067 & 0.033 & 0.034 & 0.079 & 0.147 & 0.047 & 0.036  & 0.060 \\

R3DG~\cite{gao2023relightable}    &  \cellcolor{yzysecond}0.011                    & \cellcolor{yzybest}0.037                    &   \cellcolor{yzysecond}0.015                 &  \cellcolor{yzysecond}0.007                  &   \cellcolor{yzybest}0.036                      &  \cellcolor{yzysecond}0.112                   &    \cellcolor{yzybest}0.021                  &  \cellcolor{yzythird}0.013  &  \cellcolor{yzysecond}0.032                                                                                                                                                                                                                                                                     \\

Ours & \cellcolor{yzybest}{0.010}                          & \cellcolor{yzybest}{0.037}                   & \cellcolor{yzybest}{0.014}                         & \cellcolor{yzybest}{0.006}         & \cellcolor{yzysecond}{0.037}               & \cellcolor{yzysecond}{0.112}                       & \cellcolor{yzysecond}{0.022}                           & \cellcolor{yzybest}{0.010}  &  \cellcolor{yzybest}{0.031}                                                                                                                                                                                        
 \\ \hline
\end{tabular}}
\label{tab:tab_nerfsynthetic}
\end{table*}

\begin{figure}[b]
    \centering
    % \vspace{-2ex}
    \includegraphics[width=\linewidth]{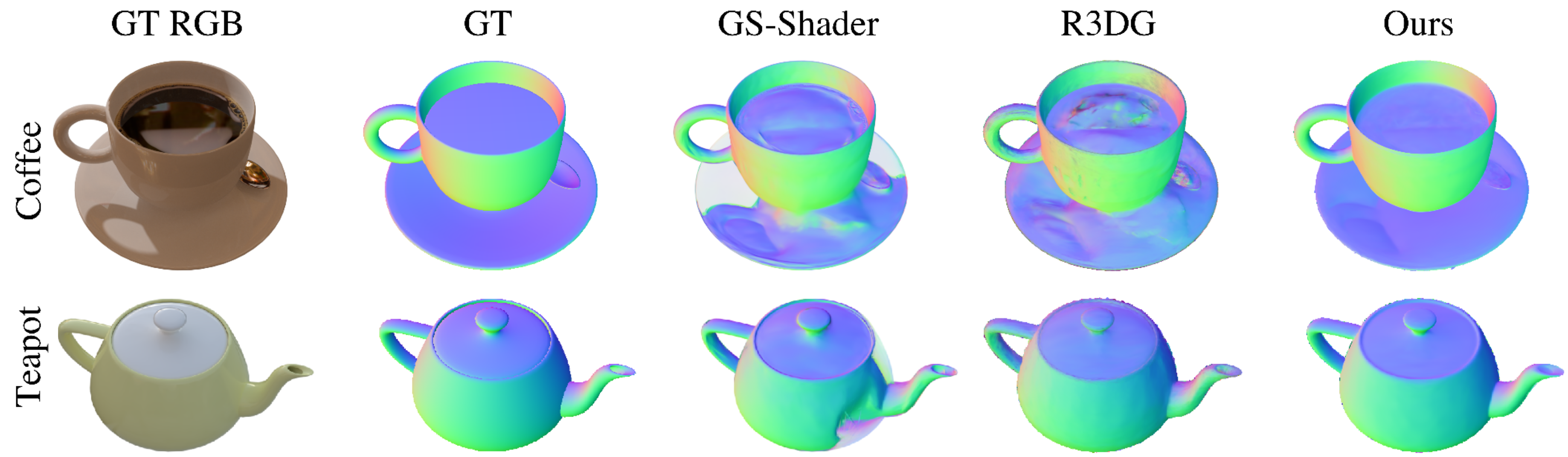}
    \caption{Visualized normal results on Shiny Blender dataset~\cite{verbin2022ref}. It illustrates the superior normal estimation achieved by our method. 
    }
    \label{fig:fig_shinyblender_normal}
          
\end{figure}

\begin{table*}[h]
\centering
\caption{Quantitative comparisons on Glossy Synthetic. We report color each cell as \colorbox{yzybest}{best}, \colorbox{yzysecond}{second best} and \colorbox{yzythird}{third best}.}
\vspace{-1ex}
\scalebox{0.9}{
\begin{tabular}{lccccccccc}
\hline
        \multicolumn{10}{c}{Glossy Synthetic~\cite{liu2023nero}}       \\
            & \multicolumn{1}{l}{Angel} & \multicolumn{1}{l}{Bell} & \multicolumn{1}{l}{Cat} & \multicolumn{1}{l}{Horse} & \multicolumn{1}{l}{Luyu} & \multicolumn{1}{l}{Potion} & \multicolumn{1}{l}{Tbell} & \multicolumn{1}{l}{Teapot} & \multicolumn{1}{l}{Avg.}           \\ \hline
\multicolumn{10}{c}{PSNR$\uparrow$}                                                                                                                                                                                                                                               \\ \hline
2D-GS~\cite{Huang2DGS2024}        & \cellcolor{yzythird}27.01                     & \cellcolor{yzythird}24.78                     & 30.68                    & 24.58                   & \cellcolor{yzysecond}26.91                        & 29.42                    & 23.50                      & 21.08   &  25.99                          \\

Scaffold-GS~\cite{lu2024scaffold}      & 26.38                     & 24.63                     & \cellcolor{yzythird}30.83                    & 24.03                  & \cellcolor{yzythird}26.90                         & \cellcolor{yzybest}30.00                    & \cellcolor{yzybest}24.77                      & \cellcolor{yzythird}21.21 & \cellcolor{yzythird}26.09                             \\

GS-Shader~\cite{jiang2024gaussianshader}    & \cellcolor{yzybest}27.19 & \cellcolor{yzybest}30.02                  & \cellcolor{yzysecond}31.34                    & \cellcolor{yzybest}26.23                   & 26.85                         & \cellcolor{yzythird}29.47                   & \cellcolor{yzysecond}23.76                      & \cellcolor{yzysecond}23.61    &  \cellcolor{yzybest}27.31                        \\

GS-IR~\cite{liang2024gs} & 23.37 & 23.40 & 28.05 & 22.75 & 25.31 & 28.05 & 18.43 & 19.76   & 23.64 \\

R3DG~\cite{gao2023relightable}    &  26.65                    & 24.56                    &   30.67                 &  \cellcolor{yzythird}24.83                  &   26.63                      &  29.06                   &    \cellcolor{yzythird}22.73                  &   20.69  &  25.73                                                                                                                                                                                                                                                                     \\

Ours & \cellcolor{yzysecond}27.07                          & \cellcolor{yzysecond}25.28                          & \cellcolor{yzybest}32.23                         & \cellcolor{yzysecond}26.03           & \cellcolor{yzybest}26.94                             & \cellcolor{yzysecond}29.48                       & 23.46                 & \cellcolor{yzybest}23.68  &  \cellcolor{yzysecond}26.77                          \\ \hline

\multicolumn{10}{c}{SSIM$\uparrow$}                                                                                                                                                                                                                                            \\ \hline
2D-GS~\cite{Huang2DGS2024}        & \cellcolor{yzythird}0.917                     & \cellcolor{yzythird}0.907                     & \cellcolor{yzythird}0.958                    & \cellcolor{yzythird}0.909                   & \cellcolor{yzysecond}0.916              & \cellcolor{yzybest}0.936                    & \cellcolor{yzybest}0.903                      & \cellcolor{yzysecond}0.881   &  \cellcolor{yzythird}0.916                          \\

Scaffold-GS~\cite{lu2024scaffold}      & 0.904                     & 0.901                     & 0.956                    & 0.888                  & 0.905                         & \cellcolor{yzythird}0.933                    & \cellcolor{yzybest}0.903                      & \cellcolor{yzythird}0.868 & 0.907                            \\

GS-Shader~\cite{jiang2024gaussianshader}    & \cellcolor{yzybest}0.924                     & \cellcolor{yzybest}0.941                     & \cellcolor{yzysecond}0.960                    & \cellcolor{yzybest}0.932                   & \cellcolor{yzythird}0.915                         & \cellcolor{yzysecond}0.935                   & \cellcolor{yzysecond}0.901                      & \cellcolor{yzybest}0.898    &  \cellcolor{yzybest}0.926                        \\

GS-IR~\cite{liang2024gs} & 0.747 & 0.796 & 0.916 & 0.748 & 0.880 & 0.895 & 0.799 & 0.834    & 0.827  \\ 

R3DG~\cite{gao2023relightable}    &  0.909                    & 0.892                    &   0.955                 &  0.904                  &   0.912                      &  0.928                   &    0.888                  &   0.867  &  0.907                                                                                                                                                                                                                                                                     \\

Ours & \cellcolor{yzysecond}0.919                          & \cellcolor{yzysecond}0.908                          & \cellcolor{yzybest}0.963                         & \cellcolor{yzysecond}0.926                        & \cellcolor{yzybest}0.919                             & \cellcolor{yzybest}0.936                       & \cellcolor{yzythird}0.900                           & \cellcolor{yzybest}0.898  &  \cellcolor{yzysecond}0.920                                                                                                                                                                                                 
 \\ \hline
 
\multicolumn{10}{c}{LPIPS$\downarrow$}                                                                                                                                                                                                                                              \\ \hline
2D-GS~\cite{Huang2DGS2024}        & \cellcolor{yzythird}0.072                     & 0.108                     & 0.060                   & \cellcolor{yzythird}0.070                   & 0.068                         & 0.097                    & \cellcolor{yzythird}0.122                      & \cellcolor{yzythird}0.101  &  \cellcolor{yzysecond}0.087                          \\

Scaffold-GS~\cite{lu2024scaffold}      & 0.081                     & \cellcolor{yzythird}0.104                     & \cellcolor{yzysecond}0.057                 & 0.083                  & 0.074                         & \cellcolor{yzysecond}0.095                    & \cellcolor{yzybest}0.111                     & 0.105 & 0.089                  \\

GS-Shader~\cite{jiang2024gaussianshader}    &\cellcolor{yzysecond}0.070                     & \cellcolor{yzybest}0.082                     & \cellcolor{yzythird}0.058                    & \cellcolor{yzybest}0.059                   & \cellcolor{yzythird}0.067                         & \cellcolor{yzythird}0.096                   & 0.129                      & \cellcolor{yzybest}0.096    &  \cellcolor{yzybest}0.082                        \\

GS-IR~\cite{liang2024gs}  & 0.113 & 0.159 & 0.104 & 0.095 & 0.087 & 0.134 & 0.204 & 0.130    & 0.128   \\

R3DG~\cite{gao2023relightable}    &  \cellcolor{yzysecond}0.070                    & 0.112                    &   0.063                 &  \cellcolor{yzysecond}0.068                  &   \cellcolor{yzysecond}0.065                      &  \cellcolor{yzybest}0.092                   &    0.125                  &   0.106  &  \cellcolor{yzythird}0.088                                                                                                                                                                                                                                                                    \\

Ours & \cellcolor{yzybest}0.067                          & \cellcolor{yzysecond}0.101                          & \cellcolor{yzybest}0.056                         & \cellcolor{yzybest}0.059                        & \cellcolor{yzybest}0.064                             & 0.097                       & \cellcolor{yzysecond}0.113                           & \cellcolor{yzybest}0.096  &  \cellcolor{yzybest}0.082                                                                                                                                                                                                 
 \\ \hline
\end{tabular}
}
\label{tab:tab_glossy}
      \vspace{-2ex}
\end{table*}

We observe that while jointly estimating all parameters, the inherent ambiguities introduced by specular lights and environment reflection complicate the differentiation between textures and lighting, leading to low-quality geometry reconstruction. Thus, we introduce phased training strategy to help decoupling the geometry and material properties. 

The training procedure is divided into three stages. In the first stage, training is guided solely by photometric losses and alpha constraints, following the approach used in 2D-GS~\cite{Huang2DGS2024}. In the second stage, normal regularization is introduced to enhance the reconstruction of robust surface geometry. In the third stage, geometry properties are fixed while optimizing only the color parameters. For specular rendering, we sample $N_s=64$ incident rays per Gaussian point. Training is conducted over 7000 iterations in the first stage, 18000 iterations in the second stage, and 15000 iterations in the final stage, totaling 40000 iterations, consistent with \cite{gao2023relightable}. All experiments are conducted on a single NVIDIA GeForce RTX 4090 GPU.

\begin{figure*}[t]
    \centering
    \vspace{-3ex}
    \includegraphics[width=\linewidth]{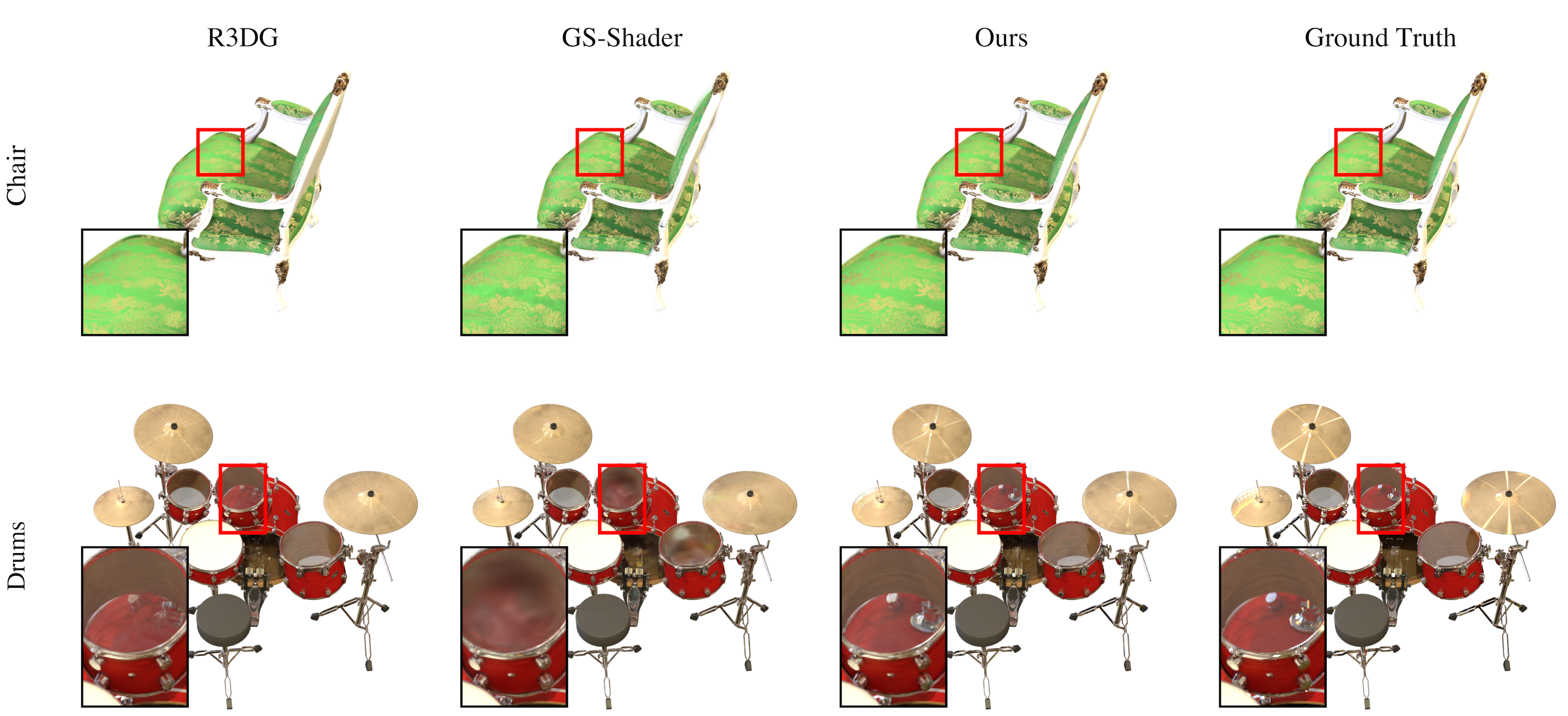}
    \caption{The qualitative comparisons on NeRF Synthetic Dataset~\cite{mildenhall2020nerf}. Our method renders the glossy surfaces with high fidelity and reconstructs accurate shadowing. Some areas are zoomed in for better visualization.
    }
    \label{fig:nerf_synthetic}
      \vspace{-3ex}
\end{figure*}

\subsection{Baselines and Metrics}
We compare our method with recent GS-based inverse rendering techniques: GaussianShader~\cite{jiang2024gaussianshader}, GS-IR~\cite{liang2024gs}, and Relightable 3DGS~\cite{gao2023relightable} that demonstrate effciency advantage over NeRF-based methods, as well as with baseline methods including 2D-GS~\cite{Huang2DGS2024} and the top-performing Scaffold-GS~\cite{lu2024scaffold} that utilizes neural-based enhancements. Quantitative metrics used include peak signal-to-noise ratio (PSNR), structural similarity index measure (SSIM), and learned perceptual image patch similarity (LPIPS).

\subsection{Comparisons}
To thoroughly demonstrate the effectiveness of GlossGau, we evaluate it on several datasets: the widely-used novel view synthesis (NVS) dataset, NeRF Synthetic~\cite{mildenhall2020nerf} and reflective object datasets: Glossy Synthetic~\cite{liu2023nero} and Shiny Blender~\cite{verbin2022ref} dataset. 

\begin{figure}
    \centering
    {\includegraphics[width=\linewidth]{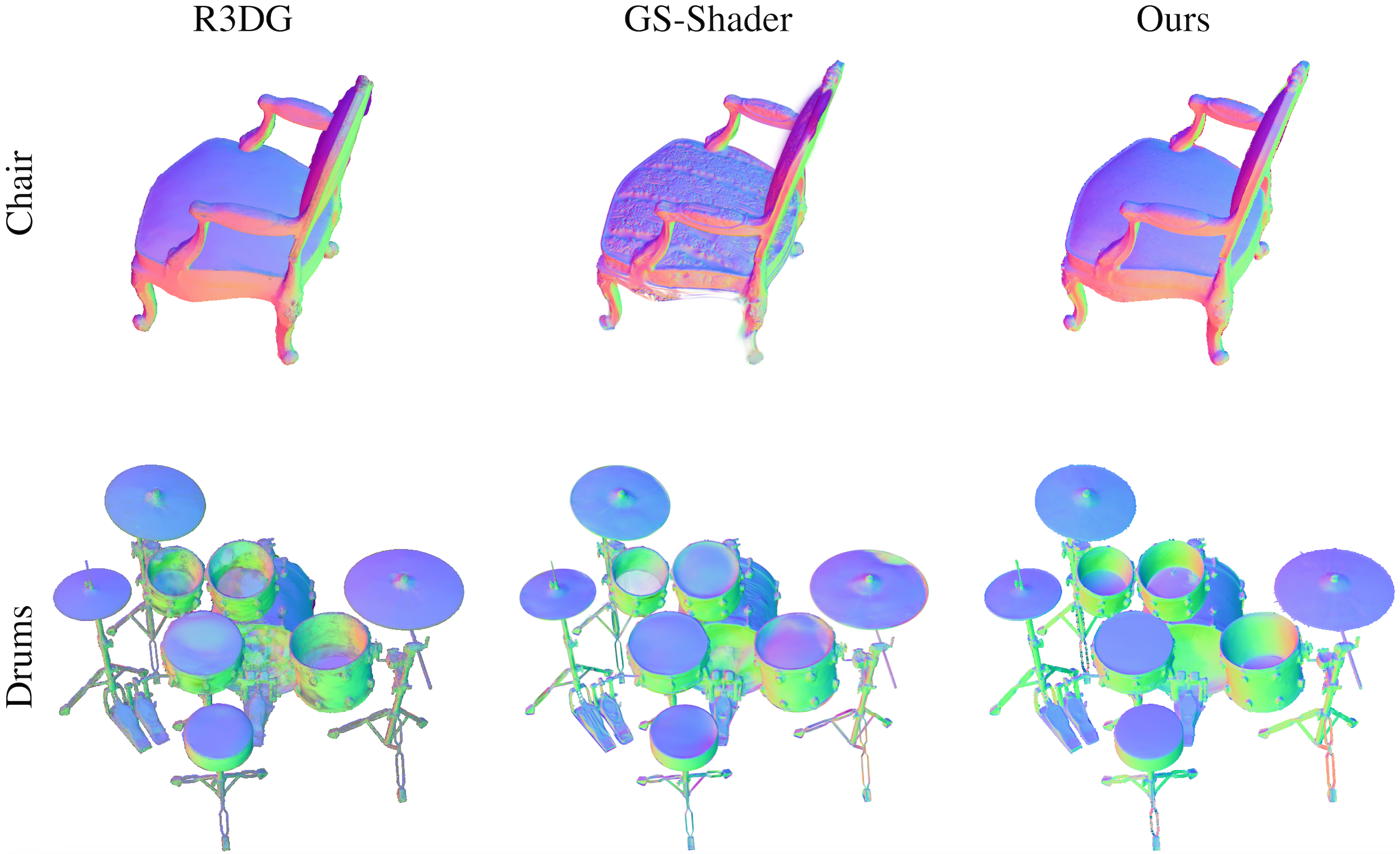}}
    \caption{Visualized normals on NeRF Synthetic Dataset~\cite{mildenhall2020nerf}.
    }
    \label{fig:nerf_normal}
    % \vspace{-3ex}
\end{figure}

\noindent{\bf NeRF Synthetic dataset.}
We first evaluate the novel view synthesis (NVS) on the NeRF synthetic dataset~\cite{mildenhall2020nerf}. The results are presented in Table \ref{tab:tab_nerfsynthetic}, and visual comparisons are shown in Figure \ref{fig:nerf_synthetic}. The corresponding predicted normals is shown in Figure \ref{fig:nerf_normal}. Our approach achieves quantitatively and perceptually comparable results with both point-based~\cite{Huang2DGS2024,lu2024scaffold} and inverse-rendering methods~\cite{jiang2024gaussianshader,gao2023relightable}. GlossGau successfully reconstructs the semi-transparent drumhead and glossy components as well as the shadowing casting of the cushion. Furthermore, GlossGau accurately disentangle the geometry and surface texture in normal estimation stage, contributing to the quality of final rendering.

\begin{figure*}[t]
    \centering
    \vspace{-3ex}
    \scalebox{0.9}{\includegraphics[width=\linewidth]{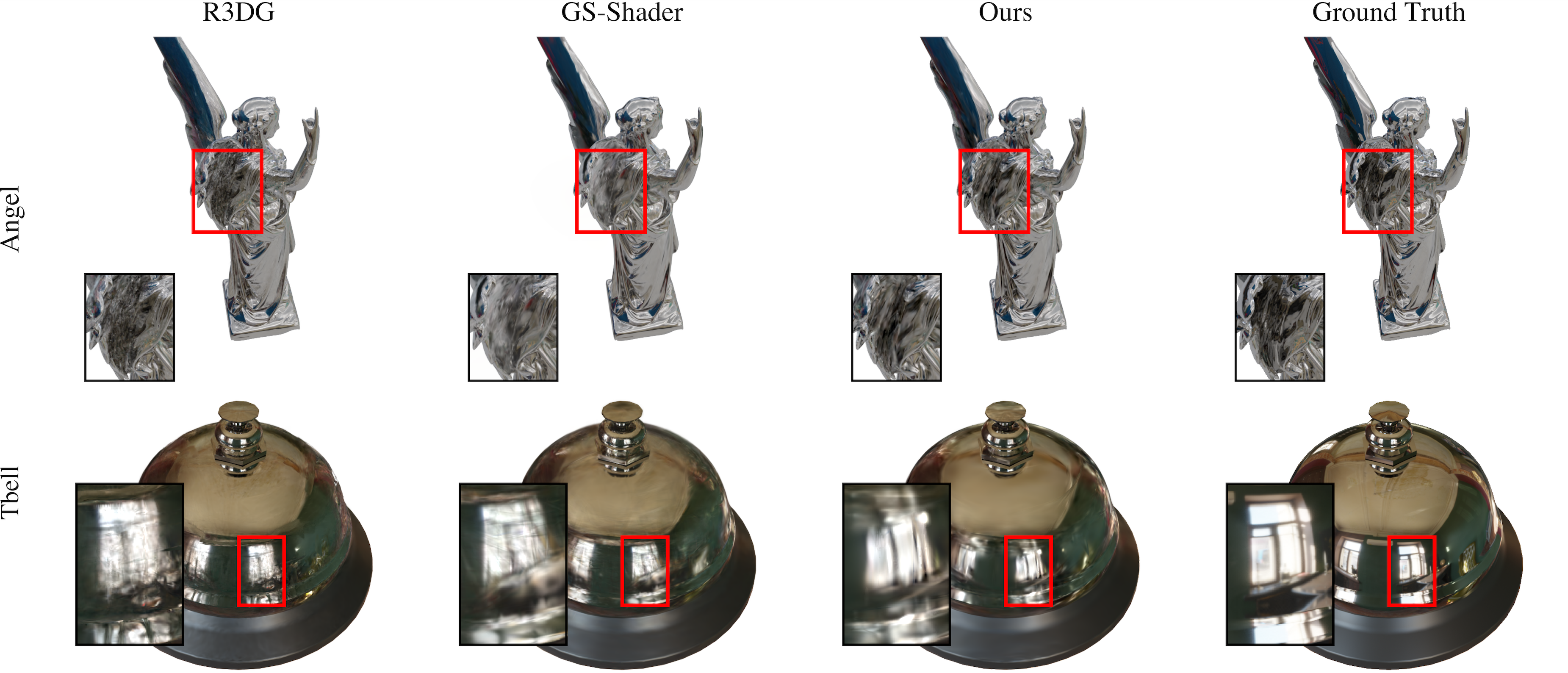}}
    \vspace{-1ex}
    \caption{The qualitative comparisons on Glossy Synthetic Dataset~\cite{liu2023nero}. Our method accurately characterizes surface roughness and generates realistic and smooth reflection effects compared with prior GS-based inverse rendering methods.
    }
    \label{fig:glossy}
      
\end{figure*}

\noindent{\bf Glossy Synthetic dataset \cite{liu2023nero}.}
Quantitative results are reported in Table \ref{tab:tab_glossy}. Our method excels across metrics and is comparable to GaussianShader~\cite{jiang2024gaussianshader}. As illustrated in Figure \ref{fig:glossy},  our method generates visually appealing rendering with less noisy surfaces and more realistic shadowing. While effectively capturing shiny regions, our approach maintains clean reconstruction in non-specular areas, yielding more photorealistic rendering results.

\begin{figure*}[t]
\vspace{-1ex}
    \centering
    \scalebox{0.9}{\includegraphics[width=\linewidth]{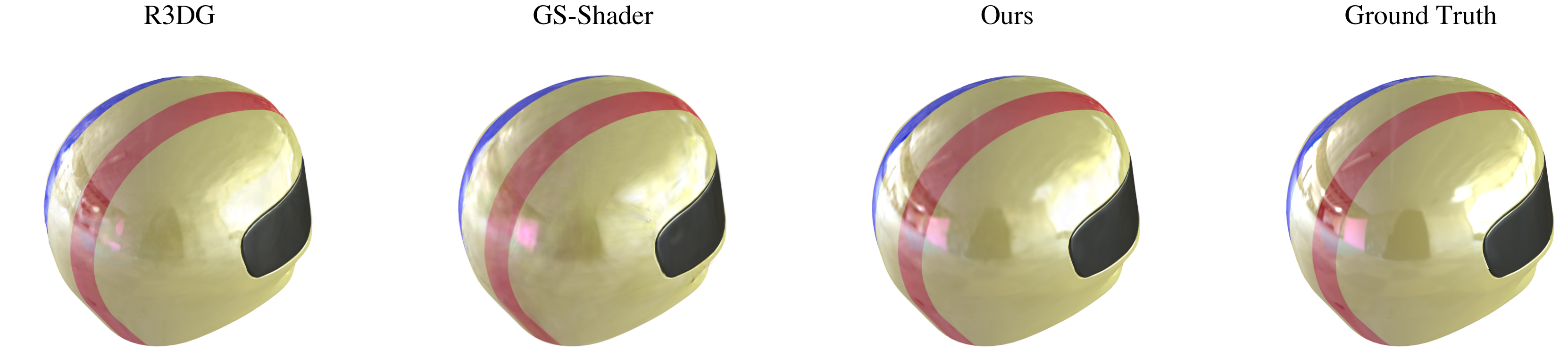}}
    \caption{Visualized comparisons on Shiny Blender Dataset~\cite{verbin2022ref}.
    }
    \label{fig:refnerf}
    \vspace{-2ex}
\end{figure*}

\noindent{\bf Shiny Blender dataset.~\cite{verbin2022ref}}
Visual comparisons on the Shiny Blender dataset are presented in Figure \ref{fig:refnerf}. Our method demonstrates superior reconstruction of glossy surfaces and accurate reflection characteristics compared to other approaches. These results validate the precision of our joint geometry and material parameter estimation. The complete results are included in the supplementary.

\begin{table}[]
\centering
\caption{Comparisons over average training time, rendering FPS, and result memory size. Our method achieves the best competitive training/rendering speed with the lowest storage consumption.}
\vspace{-1ex}
\scalebox{0.85}{
\begin{tabular}{lccc}
\hline
                     & Training Time (h)             & FPS     & Mem (MB)       \\ \hline
GS-Shader~\cite{jiang2024gaussianshader}            & 1.03 & 75 & \textbf{24.10} \\

GS-IR~\cite{liang2024gs} & 0.76 & 253 & 76.64 \\

R3DG~\cite{gao2023relightable}              & 0.73  & 9.67                & 339          \\
Ours               & \textbf{0.34} & \textbf{262}           &  29.12          \\ \hline
\end{tabular}}
\label{tab:tab_speed}
\vspace{-3ex}
\end{table}

Additionally, we report the average training time, rendering FPS, and resultant Gaussan's size in Table~\ref{tab:tab_speed}. The reported results are averaged among objects in NeRF synthetic datasets~\cite{mildenhall2020nerf}. On the same hardware, our method only takes about 0.34 hour for training and the resulting pointcloud occupies significantly less storage compared with \cite{gao2023relightable}. Given competitive or even superior reconstruction fidelity, GlossGau exhibits notable computational advantages.

\subsection{Ablation Study}
\label{sec:exp_ablation}
\begin{table}[]
\centering
\caption{Ablation studies on Normal Regularization, ASG Warping, and Phased Training on NeRF Synthetic Dataset. }
% \vspace{-1ex}
\scalebox{0.9}{
\begin{tabular}{lccc}
\hline
                                           & PSNR$\uparrow$           & SSIM$\uparrow$           & LPIPS$\downarrow$  
                                            \\ \hline
w/o $\mathcal{L}_\text{normal}$       & 28.45          & 0.910          & 0.053          \\
w/o ASG                            & 33.04          & 0.963          & 0.033           \\
w/o Phased Training                & 32.11          & 0.960          & 0.035           \\
Ours & \textbf{33.41} & \textbf{0.965} & \textbf{0.031} \\ \hline
\end{tabular}
}
\label{tab:tab_ablation}
      \vspace{-3ex}
\end{table}

We conduct ablation studies of our model on NeRF Synthetic Dataset. The numerical comparisons are provided in Table \ref{tab:tab_ablation}. We assess the effectiveness of anisotropic spherical Gaussian warping by comparing it against the standard isotropic spherical Gaussian BRDF formulation from~\cite{wang2009all}. Additionally, we ablate the normal regularization, relying solely on photometric loss for optimization. We also validate the necessity of our phased training protocol by comparing against simultaneous optimization of all Gaussian parameters. Detailed comparisons are provided in the supplementary material. 
\section{Conclusion}

We introduce GlossGau, an efficient approach to GS-based inverse rendering framework for scenes with glossy surfaces. Our method models the BRDF through singular Anisotropic Spherical Gaussian for Gaussian Splatting, which analytically approximate the microfacet normal distribution function. Additionally, we innovatively adopts surfel-based Gaussian as primitives, which combines with regularization, ensures accurate normal estimation and visibility computation with no extra cost, yielding improved decomposition of geometry and material properties. Quantitative and visual comparisons show that our method obtains competitive rendering fidelity and superior efficiency advantage over prior works.

\noindent\textbf{Acknowledgements.} This research was supported by the Ministry of Trade, Industry and Energy (MOTIE) and Korea Institute for Advancement of Technology (KIAT) through the International Cooperative R\&D program in part (P0019797). 

\newpage
{
    \small
    \bibliographystyle{ieeenat_fullname}
    \bibliography{main}
}

\end{document}